\documentclass[letterpaper]{article} 
\usepackage{aaai24}  
\usepackage{times}  
\usepackage{helvet}  
\usepackage{courier}  
\usepackage[hyphens]{url}  
\usepackage{graphicx} 
\usepackage{natbib}  
\usepackage{caption} 
\usepackage{algorithm}
\usepackage{algorithmic}
\usepackage{subcaption}
\usepackage{cuted}
\usepackage{booktabs}
\usepackage{array}
\usepackage{multirow}
\usepackage{textcomp}
\usepackage{amsmath}
\usepackage{tabularx}
\usepackage{rotating}
\usepackage{array}
\usepackage{tikz}
\usepackage{seqsplit}
\usepackage{dcolumn}

\newcolumntype{P}[1]{>{\centering\arraybackslash}m{#1}}
\newcolumntype{L}[1]{>{\raggedright\arraybackslash}p{#1}}
\newcolumntype{C}[1]{>{\centering\arraybackslash}S{m{#1}}}

\usepackage{newfloat}
\usepackage{listings}
\DeclareCaptionStyle{ruled}{labelfont=normalfont,labelsep=colon,strut=off} 
\floatstyle{ruled}
\newfloat{listing}{tb}{lst}{}
\floatname{listing}{Listing}
\title{From Artificially Real to Real: Leveraging Pseudo Data from Large Language Models for Low-Resource Molecule Discovery}

\author{
    Yuhan Chen,
    Nuwa Xi,
    Yanrui Du,
    Haochun Wang,
    Jianyu Chen,
    Sendong Zhao\thanks{Corresponding author},
    Bing Qin
}
\affiliations{
    Research Center for Social Computing and Information Retrieval\\
    Harbin Institute of Technology, China\\

    \{yuhanchen, sdzhao\}@ir.hit.edu.cn
}

\usepackage{bibentry}

\begin{document}

\maketitle

%
%
\begin{abstract}
Molecule discovery serves as a cornerstone in numerous scientific domains, fueling the development of new materials and innovative drug designs. Recent developments of in-silico molecule discovery have highlighted the promising results of cross-modal techniques, which bridge molecular structures with their descriptive annotations. However, these cross-modal methods frequently encounter the issue of data scarcity, hampering their performance and application. In this paper, we address the low-resource challenge by utilizing artificially-real data generated by Large Language Models (LLMs). We first introduce a retrieval-based prompting strategy to construct high-quality pseudo data, then explore the optimal method to effectively leverage this pseudo data. Experiments show that using pseudo data for domain adaptation outperforms all existing methods, while also requiring a smaller model scale, reduced data size and lower training cost, highlighting its efficiency. Furthermore, our method shows a sustained improvement as the volume of pseudo data increases, revealing the great potential of pseudo data in advancing low-resource cross-modal molecule discovery.
\end{abstract}

%
%
\section{Introduction}

Molecule discovery plays a critical role in numerous scientific domains including chemistry \cite{wang2023advances, cuzzucoli2023predictive}, pharmacology \cite{patani1996bioisosterism, anderson2003process}, and materials science \cite{curtarolo2013high}. However, traditional molecule design methods are frequently faced with challenges such as high costs, lengthy development processes, and limited success rates. Introducing a new drug to the market, for instance, might demand over a billion dollars and more than a decade of development \cite{gaudelet2021utilizing}.

\begin{figure}[t]
  \begin{subfigure}{0.45\textwidth}
    \centering
    \includegraphics[width=\linewidth]{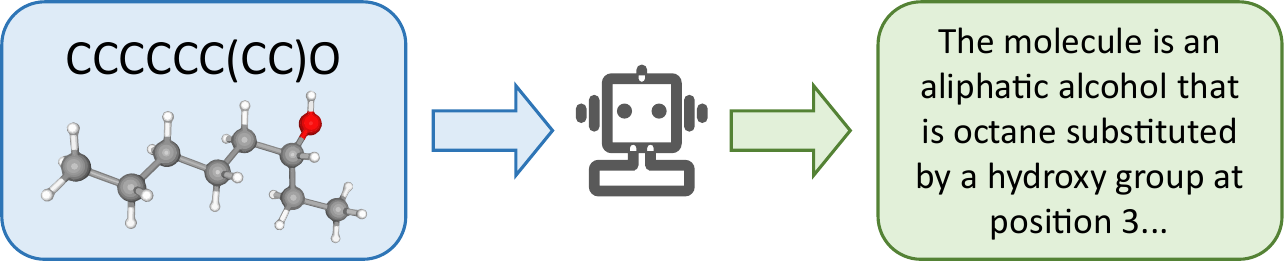}
    \caption{Molecular captioning}
    \label{fig:smi2cap}
  \end{subfigure}
  \begin{subfigure}{0.45\textwidth}
    \centering
    \includegraphics[width=\linewidth]{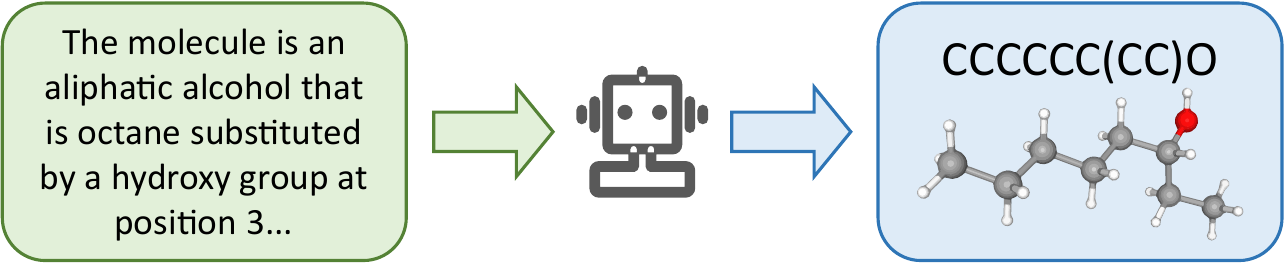}
    \caption{Text-Based de novo Molecule Generation}
    \label{fig:cap2smi}
  \end{subfigure}
  \caption{Illustration of translation between molecule and description in cross-modal molecule discovery.
  }
  \label{fig:task-demo}
\end{figure}

With the advent of artificial intelligence (AI), innovative cross-modal methods are ushering in new ways to synthesize and analyze complex molecular structures, enhancing efficiency and reshaping the fields of computational chemistry and material science. \citet{edwards2022translation} proposed a novel approach to directly translate molecules to corresponding captions and generate molecular structures from natural language text, shown in Figure \ref{fig:task-demo}. This cross-modal method heralds a future in which the design and study of specialized molecules can be achieved through simple natural language sentences.

Various attempts have been made to resolve these tasks. MolT5 \cite{edwards2022translation} uses SMILES (Simplified Molecular Input Line Entry System) \cite{weininger1988smiles} and molecule description respectively for masked language modeling (MLM) \cite{raffel2020exploring} as pre-training. \citet{liu2023molxpt} pre-train models with causal language modeling (CLM) on the sequences that blend biomedical literature with molecular structural representations, derived from replacing molecular entities with their SMILES representations. However, these studies are limited by the scarcity of parallel molecule-description pairs, rendering direct sequence-to-sequence training unfeasible. The effectiveness of sequence-to-sequence (seq2seq) training is evident in \citet{christofidellis2023unifying}, where the annotated data from the downstream dataset is incorporated for pre-training, albeit in a significantly lower ratio compared to the unannotated data. The primary bottleneck is the annotation process itself: the annotation of these pairs demands specialized knowledge in molecular chemistry, rendering large-scale human annotation both expensive and difficult.

Inspired by the great success of LLMs in natural language processing (NLP) and related fields \cite{bagal2021molgpt, frey2022neural, ferruz2022protgpt2}, we propose to mitigate the low-resource difficulty by using artificially-real data generated by LLMs. Unlike ``real data'', which originates from genuine experimental or observational sources, this ``pseudo data'' or ``artificially-real data'' is crafted artificially. While it mirrors the format of real data, its content does not depict actual real-world observations, making it potentially unsuitable for direct real-world applications.

Our approach begins by creating a comprehensive pseudo dataset intended for seq2seq pre-training.  We collect 1M unlabeled molecules from PubChem and use the in-context learning ability of LLMs to generate descriptive captions for these molecules. To ensure the integrity and diversity of this pseudo data, we adopt a retrieval-based one-shot prompting strategy during generation. Through this way, we construct the first artificially-real dataset, PseudoMD-1M, consisting of 1,020,139 pseudo molecule-description pairs.

Based on this dataset, we explore the optimal method to leverage pseudo data. We propose two primary methods: 1) using pseudo data exclusively during pre-training for domain adaptation, and 2) integrating pseudo data with real data during fine-tuning as a data augmentation technique. To offer a comprehensive evaluation, we further compile DrugBank-23, a novel dataset derived from a different data source than existing datasets.

In summary, our contributions are as follows:
\begin{itemize}
    \item We are the first to incorporate LLMs for low-resource molecule discovery. Using artificially-real data generated by LLMs, we are able to mitigate the data scarcity for the tasks. We release PseudoMD-1M, the first artificially-real dataset for cross-modal molecule discovery, which is 33$\times$ larger than existing real datasets.
    \item We explore the effective construction and utilization of pseudo data. We specifically investigate two principal techniques, including using pseudo data as domain adaptation and data augmentation. We conduct comprehensive experiments on existing datasets, and provide our new dataset called DrugBank-23, which adds a novel data source compared to current datasets.
    \item Experimental results show that despite smaller model size and amount of pre-training data, models using artificially-real data as domain adaptation outperform all prior methods. Furthermore, our method shows continuous improvement with increasing volumes of pseudo data, underscoring its promising future applications.
\end{itemize}

%
%
\section{Related Work}

\subsection{Cross-Modal Molecule Discovery}

With the advancement of in-silico molecule discovery methods, the field of molecule exploration is undergoing a transformative shift away from its resource-intensive and costly origins \cite{rifaioglu2019recent, gaudelet2021utilizing}.
\citet{edwards2021text2mol} introduce a new task Text2Mol, which uses descriptions as search queries to retrieve the target molecules. Following this, \citet{edwards2022translation} propose two innovative tasks: molecule captioning and text-guided de novo molecule generation. These tasks aim at translating between molecular structures and natural language texts. MolXPT \cite{liu2023molxpt} leverages literature annotations of molecules to construct a pre-training dataset. \citet{christofidellis2023unifying} further improves the field with multi-task learning, which combines single-domain and cross-domain datasets for joint training. Most recently, \citet{li2023empowering} propose a strategy that enables LLMs to accomplish both molecule captioning and text-guided molecule generation tasks. Here we take one step further to construct a large number of high-quality parallel data pairs, in response to the data scarcity that limits the performance of the above approaches.

\subsection{Large Language Models}

LLMs have achieved significant success in natural language processing by scaling up to billions of parameters \cite{brown2020language, ouyang2022training}. Trained on vast corpora \cite{singhal2023large}, LLMs show more general intelligence \cite{bubeck2023sparks} and remarkable capabilities such as in-context learning \cite{rubin2022learning, min2022metaicl}. They have also obtained promising performance in chemical \cite{bagal2021molgpt, frey2022neural}, biological \cite{ferruz2022protgpt2, xi2023unicorn} and medical \cite{wang2023knowledge, du2023calla} domains. Due to their great generation capability, numerous works have relied on LLMs to generate data for various purposes, including creating semantic textual similarity datasets \cite{schick2021generating}, augmenting natural language inference \cite{liu2022wanli}, automatically formulating instructions \cite{wang2022self} and improving few-shot retrieval \cite{dai2022promptagator}. Inspired by these achievements, we aim to employ LLMs to generate parallel data, addressing data scarcity in cross-modal molecule discovery.

%
%

\section{Methodology}

\subsection{Task Overview}
Here we introduce two primary tasks for cross-modal molecule discovery. First proposed by \citet{edwards2022translation}, the two tasks act as a bridge between molecule discovery and NLP and can be considered as cross-modal translation tasks.

\begin{figure*}[t]
  \centering
  \includegraphics[width=\linewidth]{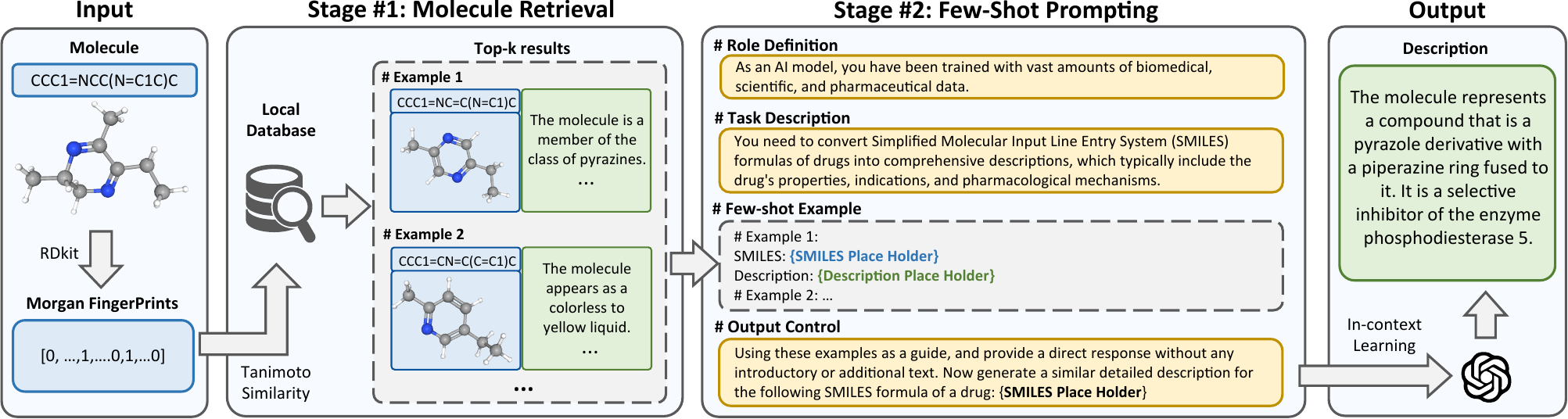}
  \caption{
  The workflow for pseudo data generation. Starting with an unlabeled molecule represented by its Morgan Fingerprints, two stages are involved. In stage 1, the input molecule serves as a search query to retrieve the top-k similar molecules from a local database containing 37,898 annotated molecule-caption pairs. In stage 2, the retrieved molecules and their captions are integrated into a prompt. Then LLMs perform in-context learning and generate a description for the input molecule.
  }
  \label{fig:data-generation}
\end{figure*}

\subsubsection{Molecular Captioning}
As illustrated in Figure \ref{fig:smi2cap}, 
Given the SMILES representation $\mathcal{S}_\mathcal{M}$ of molecule $\mathcal{M}$, the task is to generate the corresponding descriptions $\mathcal{D}_\mathcal{M}$.

\subsubsection{Text-Based De Novo Molecule Generation}

As shown in Figure \ref{fig:cap2smi}, given the descriptions $\mathcal{D}_\mathcal{M}$ of molecules $\mathcal{M}$, the task is to generate its corresponding SMILES $\mathcal{S}_\mathcal{M}$. 

\subsection{Artificially-Real Data Generation}
High-quality pseudo data is the foundation for further exploration. Here we propose PseudoMD-1M, the first pseudo dataset composed of 1M parallel molecule-description data pairs. To acquire sufficient data, we leverage a vast number of unlabeled molecules and use LLMs to generate corresponding descriptions. We begin by collecting 1.1 million unannotated SMILES strings of molecules from PubChem \cite{kim2023pubchem}. We then employ a rigorous filtering procedure to filter out the SMILES in downstream datasets to ensure that there is no overlap between the collected molecules and those contained in the real datasets \cite{edwards2021text2mol, zeng2022deep}. By doing so, we ensure that no supplementary information about the molecules present in the real datasets is accidentally incorporated, thereby maintaining the integrity and independence of the training process. With ChatGPT API, we generate textual descriptions that encompass key aspects such as properties and structural features for each unannotated molecule. To improve the quality of generated descriptions, we implement a retrieval-based prompt paradigm that comprises two main stages as follows: Molecule Retrieval and Few-Shot Prompting. 

\begin{figure}[t!]
  \centering
  \includegraphics[width=\linewidth]{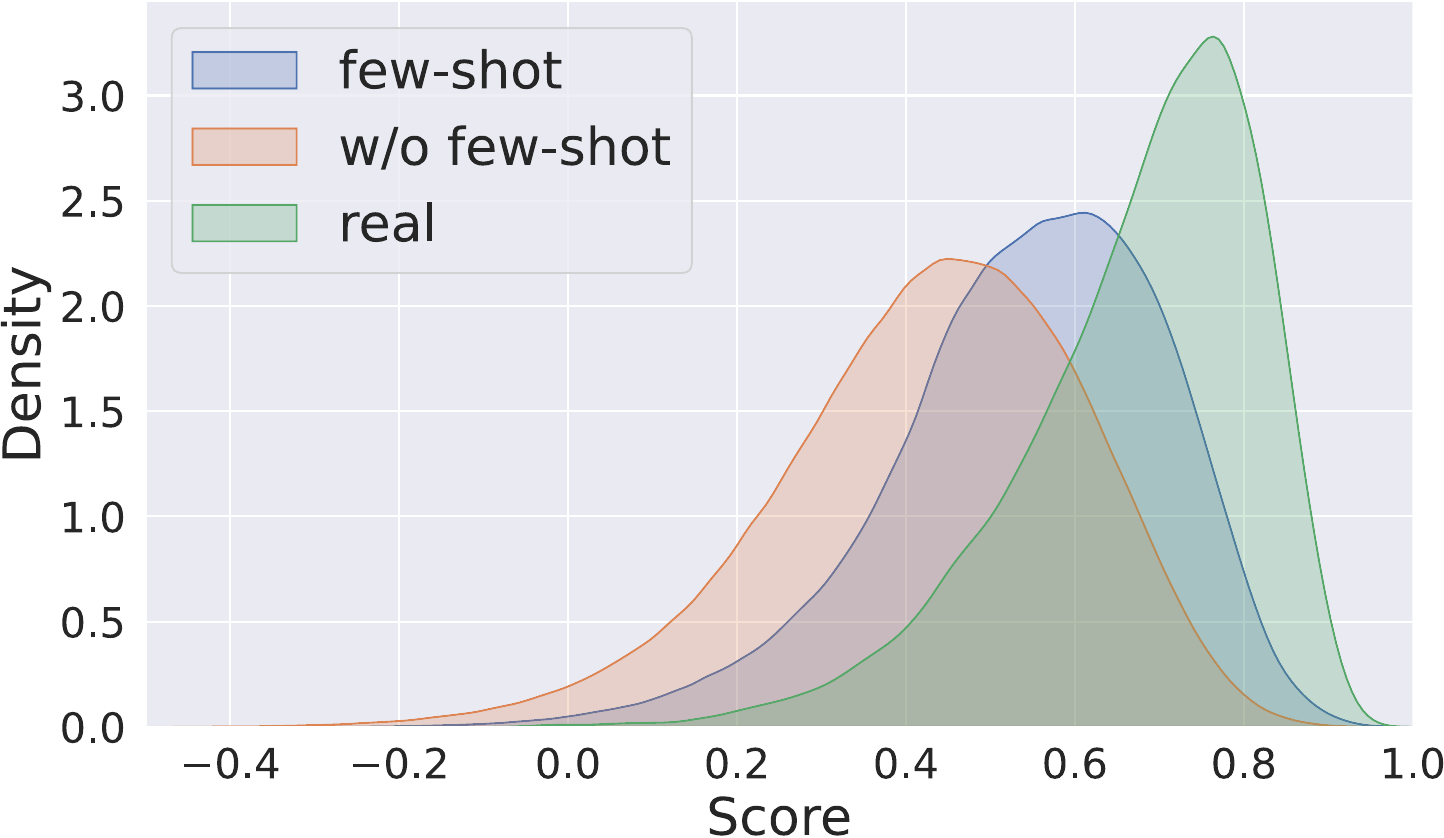}
  \caption{
  Comparison of data quality. We use the method proposed by \citet{edwards2022translation} to evaluate the similarity between molecule-description pairs as an estimation of the data quality. The distribution is visualized using Kernel Distribution Estimation. A higher Text2Mol score signifies closer molecule-description resemblance, and ``Density" represents the data concentration in a given region.
  }
  \label{fig:data-quality}
\end{figure}

\begin{figure*}[th]
  \centering
  \includegraphics[width=0.88\linewidth]{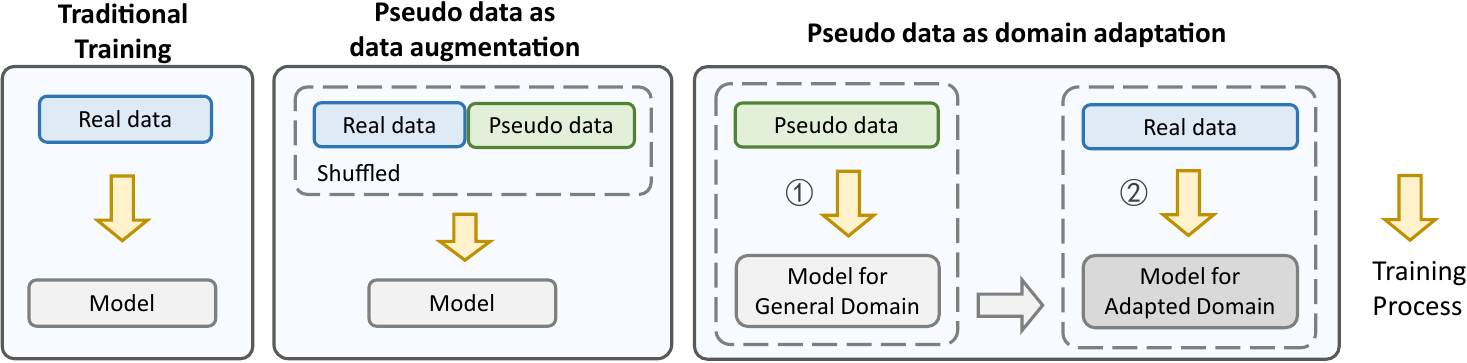}
  \caption{Different methods for utilizing pseudo data. Traditional training employs only the real dataset for fine-tuning. The data augmentation approach fine-tunes the model on the combined dataset with pseudo data incorporated. In the domain adaptation method, the model is (1) initially pre-trained on two concurrent cross-modal translation tasks using pseudo data as domain adaptation, and (2) further trained on each task using real data.}
  \label{fig:model}
\end{figure*}

\subsubsection{Molecule Retrieval}
In-context learning \cite{brown2020language} is one of the emergent abilities of LLMs, and the instances used in the prompts given to the LLMs play an important role in the generation quality. As molecules with similar structures often display corresponding characteristics \cite{wang2016improving}, we retrieve the descriptions of annotated molecules that resemble the unlabeled molecule, using them as the few-shot instance during prompting. Specifically, we collect 37,898 annotated molecules with captions from PubChem\cite{kim2023pubchem}, then retrieve the molecules with top-k Tanimoto similarity \cite{tanimoto1958elementary}, a standard measure in cheminformatics. To prevent information leakage during testing, we exclude the molecules that are contained in the real data test set \cite{edwards2021text2mol, zeng2022deep}. This process enables the models to learn from the information embedded within the descriptions of molecules that possess similar properties, ensuring a more tailored and accurate representation. Figure \ref{fig:data-quality} shows the estimate of the data quality, indicating that the few-shot prompting approach (in blue) yields higher-quality data, more closely resembling real data than without.

\subsubsection{Few-Shot Prompting}
Upon retrieving the top-k results for each unlabeled molecule from our local database, we select one example using a weighted distribution, where molecules with higher similarity have a greater chance of being chosen. This selected example is then incorporated into the final prompt. We opt for one-shot prompting to minimize generation costs, as expenses increase linearly with the number of instances included in few-shot prompts. This weighted selection method prevents repetitive selection of the same molecule as the few-shot example, thereby improving the diversity during generation while maintaining the similarity between the molecule to be annotated and the few-shot example. 
As shown in Figure \ref{fig:data-generation}, the complete prompt comprises role definition, task description, few-shot example, and output control. The role definition and task description give LLMs the general context and enable its learned knowledge, while the few-shot example acts like a supplementary material for the LLMs to refer to. Then, with the output control for format clarification, the LLMs should be able to generate the desired description. 

\subsection{Approaches to Utilize Artificially Real Data}
The ways to utilize the pseudo data decide how the model will perform on real data. We propose and explore two primary strategies to optimize the use of pseudo data.

\subsubsection{Pseudo Data as Data Augmentation}

Data augmentation strategy can be roughly categorized into two kinds, modification of existing data and generation of pseudo data. The former takes an existing data instance and makes certain alterations to it without changing its inherent meaning or label, such as rotation, flipping, and cropping for images \cite{krizhevsky2012imagenet}, or synonym replacement for text \cite{wang2015s, wei2019eda, miao2020snippext}. This method is more about adding variability and noise to existing data instances than generating completely new ones. The latter, on the other hand, involves creating new data instances that did not exist in the original dataset based on the characteristics and distribution of the original data, which is an efficient alternative when real data is scarce or when creating new real data is costly or unfeasible. Existing applications include back translation for text \cite{sennrich2016improving}, and GANs for images \cite{goodfellow2014generative}. 

Inspired by the latter techniques, we explore the use of pseudo data as data augmentation. As shown in Figure \ref{fig:model}, we keep the original data in the training set and augment them with pseudo data during fine-tuning. Using the same method as described in Figure \ref{fig:data-quality}, we assess the distribution of the real training set and the sample the augmented pseudo data based on the same distribution, ensuring consistency in the overall dataset distribution before and after data augmentation. We hope that this data augmentation approach using pseudo data will expose the model to a broader range of data patterns and scenarios, thus enhancing its ability to recognize complex patterns and generalize its learning to unseen data.

\subsubsection{Pseudo Data as Domain Adaptation}
Models pre-trained on general domain might perform less ideally when it is applied to specific domains for which they were not explicitly trained \cite{malte2019evolution}. In our case, the SMILES appears as an unfamiliar symbol to such models, making the direct fine-tuning approach less efficient. To bridge this gap, we use pseudo data as a second pre-training stage for domain adaptation. As shown in Figure \ref{fig:model}, we train the model using pseudo data for two concurrent cross-modal translation tasks: molecular captioning and text-based de novo molecule generation. Using a direct and bidirectional seq2seq approach, this stage is intended to empower the model to not only recognize the SMILES representation but also to grasp the relationship between natural language and SMILES. Given that our primary focus at this stage is not on data authenticity, pseudo data emerges as a preferable choice, particularly because it provides a large number of parallel data pairs for supervised seq2seq training compared to real datasets. We then further fine-tune it on real data to refine and enhance the model's understanding of SMILES for further authenticity -- a critical aspect for applications like drug discovery.

%
%
\section{Experiments}
To validate the effectiveness of using pseudo data, we conduct comprehensive experiments comparing our proposed approaches with existing methods. We further conduct experiments to demonstrate how the balance between real data and pseudo data could affect model performance. 
All the experiments are conducted on both molecular captioning and molecule generation. 
The implementation details are listed in Appendix C.

\subsection{Settings}
\subsubsection{Datasets}
Currently, only a few datasets with parallel molecule-description pairs exist, including ChEBI-20 \cite{edwards2021text2mol} and PCdes \cite{zeng2022deep}, both constructed using data from PubChem \cite{kim2023pubchem}. To enhance evaluation comprehensiveness, we assemble a new dataset called DrugBank-23, based on DrugBank \cite{wishart2018drugbank}. We experiment on all three datasets (ChEBI-20, PCdes, and DrugBank-23). The detailed information about these datasets is listed in Table \ref{tab:datasets}.

\begin{table}[t]
\renewcommand{\arraystretch}{1.15}
  \centering
    \begin{tabular}{c|ccc}
    \hline
    \hline
    Info  & ChEBI-20 & PCdes & DrugBank-23 \\
    \hline
    Train & 26,407 & 10,500 & 17,109 \\
    Validation & 3,301 & 1,500 & 3,667 \\
    Test  & 3,300 & 3,000 & 3,666 \\
    $\mathcal{L}_{\text{SMILES}}$ & 81.56 & 56.47 & 54.11 \\
    $\mathcal{L}_{\text{Description}}$ & 52.88 & 72.47 & 65.04 \\
    Data source & \multicolumn{2}{c}{PubChem} & DrugBank \\
    \hline
    \hline
    \end{tabular}%
    \caption{Details about the existing datasets and ours (DrugBank-23). $\mathcal{L}_{\text{SMILES}}$ denotes the average length of SMILES while $\mathcal{L}_{\text{Description}}$ denotes the average word count per description.}
  \label{tab:datasets}%
\end{table}%

\subsubsection{Models}
We evaluate the following methods:

\begin{itemize}
\item \textbf{T5} \cite{raffel2020exploring}. T5 directly fine-tuned on downstream datasets.

\item\textbf{MolT5} \cite{edwards2022translation}. T5 pre-trained with MLM using SMILES and molecule descriptions respectively, then fine-tuned on downstream datasets.
 
\item\textbf{ChatGPT} \cite{li2023empowering}. \texttt{GPT-3.5-Turbo} using few-shot prompting strategy. We cite the results from the original paper on ChEBI-20, then apply the same strategy to test on the other datasets.

\item\textbf{MolXPT} \cite{liu2023molxpt}. GPT-2 pre-trained with CLM using abstracts of biomedical literature where molecules are replaced with the corresponding SMILES, then fine-tuned on downstream datasets. As the model is currently unavailable, we cite their results on ChEBI-20.

\item\textbf{Text\&Chem T5} \cite{christofidellis2023unifying}. T5 pre-trained using multi-task learning, then fine-tuned on downstream datasets.

\item\textbf{Aug-T5} (ours). T5 fine-tuned on datasets augmented with pseudo data from PseudoMD-1M, sampled from 1k to 512k, doubling at each step. We report the optimal performances for each dataset. See Appendix D for details.

\item\textbf{Ada-T5} (ours). T5 pre-trained using molecule-description pairs from PseudoMD-1M as domain adaptation, then fine-tuned on downstream datasets.

\end{itemize}

\begin{table}[t]
\renewcommand{\arraystretch}{1.15}
  \centering
    \begin{tabular}{c|ccc}
    \hline
    \hline
    Model & Data scale & Steps & Backbone\\
    \hline
    MolT5 & 500M  & 1M  & T5\textsubscript{large} \\
    MolXPT & 8M    & 200k  & GPT2\textsubscript{medium}\\
    Text\&Chem T5 & 33.5M & 131k  & T5\textsubscript{base}\\
    Aug-T5 & 0     & 0   & T5\textsubscript{small}\\
    Ada-T5 & 1M    & 100k  & T5\textsubscript{small}\\
    \hline
    \hline
    \end{tabular}%
    \caption{Pre-training details for different Models. “M” stands for million and “k” denotes thousand.}
  \label{tab:pretrain}%
\end{table}%

As shown in Table \ref{tab:pretrain}, both our proposed methods utilize the smallest model scale, pre-training data, and steps, while Aug-T5 requires no additional pre-training. We first test our methods on T5\textsubscript{small} (Aug-T5/Ada-T5) and then apply them to T5\textsubscript{base} (Aug-T5\textsubscript{base}/Ada-T5\textsubscript{base}).

\begin{table*}[th]
\renewcommand{\arraystretch}{1.15}
\setlength{\tabcolsep}{5.5pt}
  \centering
  \begin{tabular*}{\textwidth}{c|c|lll|lll|lll}
    \hline
    \hline
    \multirow{2}{*}{Model} & \multirow{2}{*}{Parameters} & \multicolumn{3}{c|}{ChEBI-20} & \multicolumn{3}{c|}{PCdes} & \multicolumn{3}{c}{DrugBank-23} \\
          &       & 
\multicolumn{1}{c}{BL}  & \multicolumn{1}{c}{RG}  & \multicolumn{1}{c|}{MET}   & \multicolumn{1}{c}{BL}  & \multicolumn{1}{c}{RG}  & \multicolumn{1}{c|}{MET}   & \multicolumn{1}{c}{BL}  & \multicolumn{1}{c}{RG}  & \multicolumn{1}{c}{MET} \\
    \hline
    T5    & 800M  & 0.467$^\dagger$$^*$ & 0.478$^\dagger$$^*$ & 0.586$^\dagger$$^*$ & 0.252$^\dagger$$^*$ & 0.259$^\dagger$$^*$ & 0.367$^\dagger$$^*$ & 0.272$^\dagger$$^*$ & 0.299$^\dagger$$^*$ & 0.396$^\dagger$$^*$ \\
    MolT5 & 800M  & 0.508$^\dagger$ & 0.510$^\dagger$$^*$ & 0.614$^\dagger$ & 0.266$^\dagger$ & 0.272$^\dagger$ & 0.380$^\dagger$$^*$ & 0.293$^\dagger$ & 0.317$^\dagger$ & 0.416$^\dagger$ \\
    MolXPT & 350M  & 0.505$^\dagger$$^*$ & 0.511$^\dagger$$^*$ & 0.626$^\dagger$ & \multicolumn{1}{c}{-}     & \multicolumn{1}{c}{-}     & \multicolumn{1}{c|}{-}     & \multicolumn{1}{c}{-}     & \multicolumn{1}{c}{-}     & \multicolumn{1}{c}{-} \\
    Text\&Chem T5 & 250M  & 0.542$^\dagger$ & 0.543$^\dagger$ & 0.648$^\dagger$ & 0.266$^\dagger$ & 0.274$^\dagger$ & 0.382$^\dagger$ & 0.280$^\dagger$$^*$ & 0.312$^\dagger$$^*$ & 0.413$^\dagger$$^*$ \\
    ChatGPT & -     & 0.482$^\dagger$$^*$ & 0.450$^\dagger$$^*$ & 0.585$^\dagger$$^*$ & 0.194$^\dagger$$^*$ & 0.193$^\dagger$$^*$ & 0.315$^\dagger$$^*$ & 0.191$^\dagger$$^*$ & 0.218$^\dagger$$^*$ & 0.325$^\dagger$$^*$ \\
    \hline
    Aug-T5 & 77M   & 0.515 & 0.517 & 0.621 & 0.270 & 0.275 & 0.385 & 0.297  & 0.322  & 0.421  \\
    Aug-T5\textsubscript{base} & 250M   & 0.516  & 0.520  & 0.620  & 0.268  & 0.272  & 0.383  & 0.294  & 0.316  & 0.416  \\
    Ada-T5 & 77M   & 0.553 & 0.552 & 0.652 &  \textbf{0.295} & 0.295 & 0.406 & 0.310 & 0.337 & 0.435 \\
    Ada-T5\textsubscript{base} & 250M  & \textbf{0.564} &  \textbf{0.562} &  \textbf{0.660} &  \textbf{0.295} &  \textbf{0.297} &  \textbf{0.409} &  \textbf{0.322} &  \textbf{0.346} &  \textbf{0.445} \\
    \hline
    \hline
    \end{tabular*}%
    \caption{Results of different models for molecular captioning
    on ChEBI-20, PCdes and DrugBank-23 datasets. $^\dagger/^*$ denotes that Ada-T5\textsubscript{base}/Aug-T5\textsubscript{base} perform significantly better than baselines at $p\mathrm{-value}<0.01$ using t-test. The best scores are in bold. BL: BLEU-4. RG: ROUGE-2. MET: METEOR.}
  \label{tab:smi2cap}%
\end{table*}%

\begin{table*}[th]
\renewcommand{\arraystretch}{1.15}
\setlength{\tabcolsep}{5.5pt}
  \centering
  \begin{tabular*}{\textwidth}{c|c|lll|lll|lll}
    \hline
    \hline
    \multirow{2}{*}{Model} & \multirow{2}{*}{Parameters} & \multicolumn{3}{c|}{ChEBI-20} & \multicolumn{3}{c|}{PCdes} & \multicolumn{3}{c}{DrugBank-23} \\
          &       & 
\multicolumn{1}{c}{Acc}  & \multicolumn{1}{c}{Val}  & \multicolumn{1}{c|}{MAC}   & \multicolumn{1}{c}{Acc}  & \multicolumn{1}{c}{Val}  & \multicolumn{1}{c|}{MAC}   & \multicolumn{1}{c}{Acc}  & \multicolumn{1}{c}{Val}  & \multicolumn{1}{c}{MAC} \\
    \hline
    T5    & 800M  & 0.279$^\dagger$$^*$ & 0.902$^\dagger$$^*$ & 0.823$^\dagger$$^*$ & 0.089$^\dagger$ & 0.910$^\dagger$$^*$ & 0.698$^\dagger$ & 0.131$^\dagger$$^*$ & 0.923$^\dagger$$^*$ & 0.682$^\dagger$ \\
    MolT5 & 800M  & 0.311$^\dagger$$^*$ & 0.905$^\dagger$$^*$ & 0.834$^\dagger$$^*$ & 0.097$^\dagger$ & 0.925$^\dagger$ & 0.695$^\dagger$ & 0.145$^\dagger$$^*$ & 0.947$^\dagger$ & 0.686$^\dagger$ \\
    MolXPT & 350M  & 0.215$^\dagger$$^*$ &  \textbf{0.983} & 0.859$^\dagger$$^*$ & \multicolumn{1}{c}{-}     & \multicolumn{1}{c}{-}     & \multicolumn{1}{c|}{-}     & \multicolumn{1}{c}{-}     & \multicolumn{1}{c}{-}     & \multicolumn{1}{c}{-} \\
    Text\&Chem T5 & 250M  & 0.322$^\dagger$$^*$ & 0.943$^\dagger$$^*$ & 0.901$^\dagger$ & 0.105$^\dagger$ & 0.849$^\dagger$$^*$ & 0.697$^\dagger$ & 0.149$^\dagger$ & 0.898$^\dagger$$^*$ &  0.705 \\
    ChatGPT & -     & 0.139$^\dagger$$^*$ & 0.887$^\dagger$$^*$ & 0.847$^\dagger$$^*$ & 0.044$^\dagger$$^*$ & 0.867$^\dagger$$^*$ & 0.671$^\dagger$$^*$ & 0.048$^\dagger$$^*$ & 0.852$^\dagger$$^*$ & 0.665$^\dagger$$^*$  \\
    \hline
    Aug-T5 & 77M   & 0.305  & 0.907  & 0.877  & 0.070 & 0.892 & 0.700 & 0.141  & 0.911  & 0.685  \\
    Aug-T5\textsubscript{base} & 250M   & 0.386  & 0.955  & 0.884  & 0.098 & 0.927 & 0.696 & 0.158  & 0.952  & 0.681  \\
    Ada-T5 & 77M   & 0.449 & 0.967 & 0.905 & 0.135 & 0.945 & 0.725 & 0.170 & 0.955 & 0.696 \\
    Ada-T5\textsubscript{base} &  250M  &  \textbf{0.486} & 0.974 &  \textbf{0.911} &  \textbf{0.150} &  \textbf{0.956} &  \textbf{0.743} &  \textbf{0.192} &  \textbf{0.969} &  \textbf{0.706} \\
    \hline
    \hline
    \end{tabular*}%
    \caption{Results of different models for molecule generation on ChEBI-20, PCdes and DrugBank-23 datasets. $^\dagger/^*$ denotes that Ada-T5\textsubscript{base}/Aug-T5\textsubscript{base} perform significantly better than baselines at $p\mathrm{-value}<0.01$ using t-test. The best scores are in bold. Acc: Accuracy. Val: Validity. MAC: MACCS FTS.}
  \label{tab:cap2smi}%
\end{table*}%

\begin{figure*}[ht]
\centering
  \begin{subfigure}{0.23\textwidth}
    \centering
    \includegraphics[width=\linewidth]{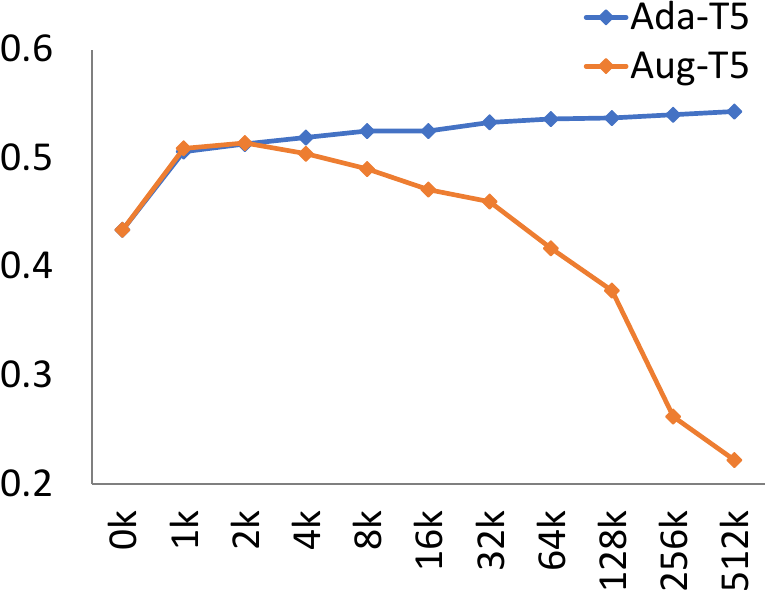}
    \caption{BLEU-4}
    \label{fig:smi2cap_bleu4}
  \end{subfigure}
  \begin{subfigure}{0.23\textwidth}
    \centering
    \includegraphics[width=\linewidth]{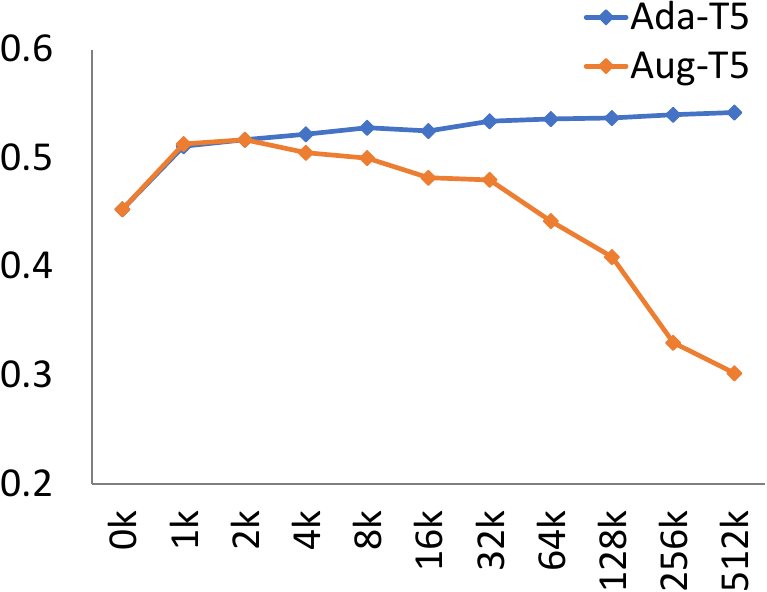}
    \caption{ROUGE-2}
    \label{fig:smi2cap_rouge2}
  \end{subfigure}
  \begin{subfigure}{0.23\textwidth}
    \centering
    \includegraphics[width=\linewidth]{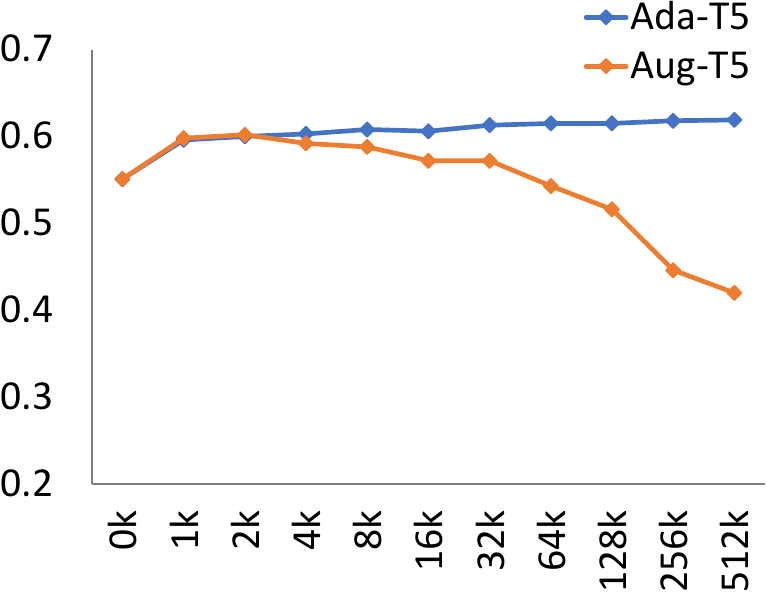}
    \caption{ROUGE-L}
    \label{fig:smi2cap_rougel}
  \end{subfigure}
  \begin{subfigure}{0.23\textwidth}
    \centering
    \includegraphics[width=\linewidth]{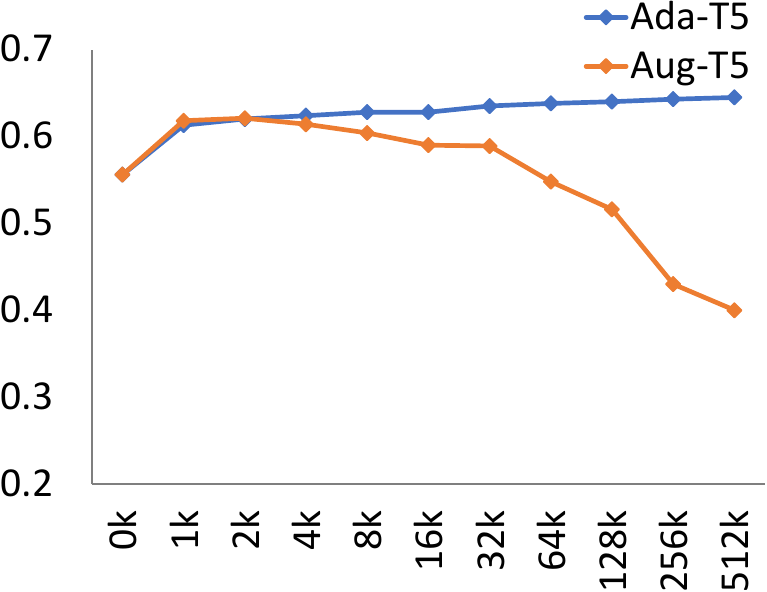}
    \caption{METEOR}
    \label{fig:smi2cap_meteor}
  \end{subfigure}
  \caption{Results of molecular captioning task using different amount of pseudo data.}
  \label{fig:ada_aug_smi2cap}
\end{figure*}

\begin{figure*}[ht]
\centering
  \begin{subfigure}{0.23\textwidth}
    \centering
    \includegraphics[width=\linewidth]{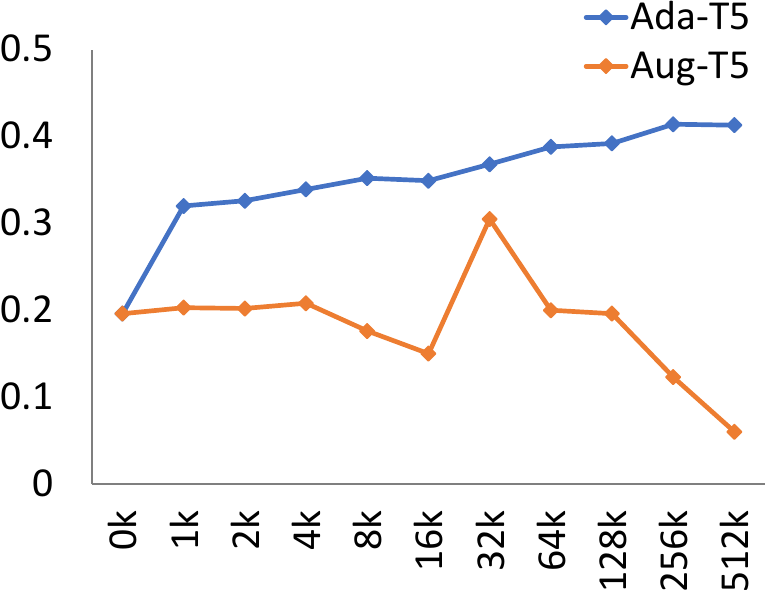}
    \caption{Accuracy}
    \label{fig:cap2smi_acc}
  \end{subfigure}
  \begin{subfigure}{0.23\textwidth}
    \centering
    \includegraphics[width=\linewidth]{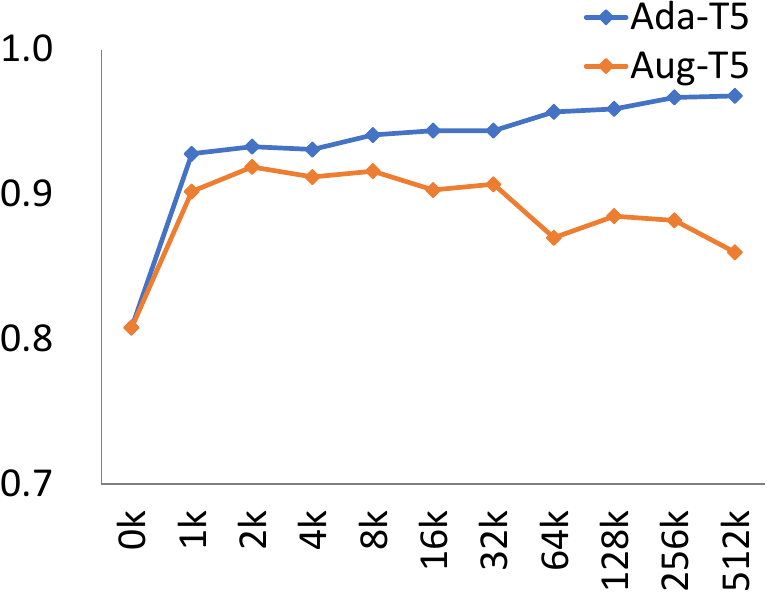}
    \caption{Validity}
    \label{fig:cap2smi_validty}
  \end{subfigure}
  \begin{subfigure}{0.23\textwidth}
    \centering
    \includegraphics[width=\linewidth]{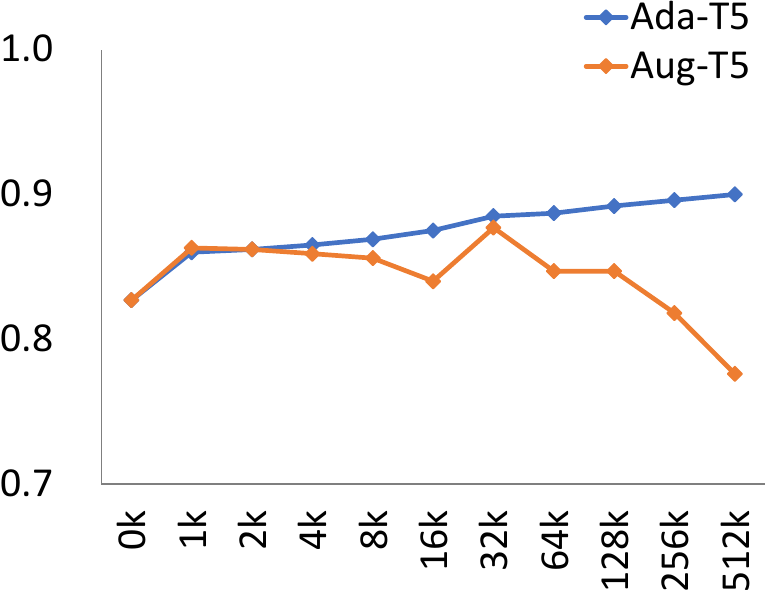}
    \caption{MACCS FTS}
    \label{fig:cap2smi_maccs}
  \end{subfigure}
  \begin{subfigure}{0.23\textwidth}
    \centering
    \includegraphics[width=\linewidth]{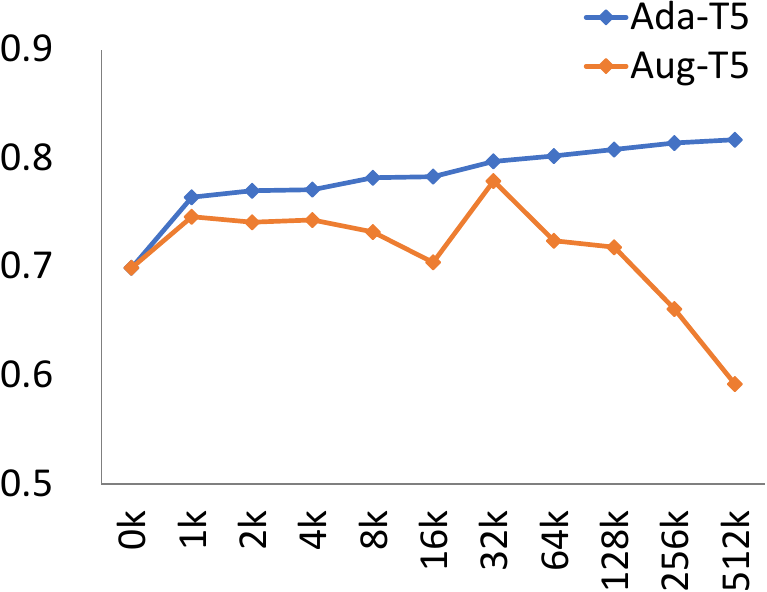}
    \caption{RDK FTS}
    \label{fig:cap2smi_rdk}
  \end{subfigure}
  \caption{Results of molecule generation task using different amount of pseudo data.}
  \label{fig:ada_aug_cap2smi}
\end{figure*}

\subsubsection{Metrics}

Following existing studies \cite{edwards2022translation, liu2023molxpt, christofidellis2023unifying}, we evaluate the results for molecular captioning with BLEU-2, BLEU-4 \cite{papineni2002bleu}, ROUGE-1, ROUGE-2, ROUGE-L \cite{lin2004rouge} and METEOR \cite{banerjee2005meteor}, and BLEU-4 \cite{papineni2002bleu}, Accuracy \cite{edwards2022translation}, Validity \cite{polykovskiy2020molecular}, Levenshtein distance \cite{miller2009levenshtein}, MACCS-FTS \cite{durant2002reoptimization}, RDK-FTS \cite{schneider2015get}, Morgan-FTS \cite{rogers2010extended} and FCD \cite{preuer2018frechet} for text-based de novo molecule generation. Selected metrics are presented in Tables \ref{tab:smi2cap}, \ref{tab:cap2smi} and Figures \ref{fig:ada_aug_smi2cap} and \ref{fig:ada_aug_cap2smi}, with comprehensive results in Appendix D.

\subsection{Comparison with Existing Methods}

\subsubsection{Results on Molecular Captioning}
Table \ref{tab:smi2cap} shows the results of different models for molecule captioning. Ada-T5 outperforms all previous methods and achieves the state-of-the-art on all three datasets across all the metrics. 
Compared to the previous state-of-the-art, Ada-T5 uses less than 3\% of the pre-training data and only a third of the model parameters, yet requires fewer training steps, demonstrating the effectiveness and computational efficiency of high-quality pseudo data. On the other hand, Aug-T5 outperforms T5, MolT5, ChatGPT and has comparable performance with MolXPT and Text\&Chem T5, using 9\%-30\% of the parameters and requires no pre-training. This highlights the benefit from the enhanced diversity of descriptions by incorporating pseudo data into the training set.
Meanwhile, Ada-T5\textsubscript{base} makes an extra but relatively little progress compared to Ada-T5, indicating that although using pseudo data for domain adaptation could also benefit from the expansion of model size like most methods, the exploitation of pseudo data only demands a relatively small number of parameters. In contrast, Aug-T5\textsubscript{base} mirrors the results of its smaller version, indicating that for data augmentation, simply increasing the model scale may not offer substantial benefits. One thing to notice is that despite the data used to train the model is generated by ChatGPT API, both our trained models can still beat ChatGPT across different metrics. This indicates that although ChatGPT can accomplish the task to a certain extent, the data it generated can still help the models achieve a more seamless transition through pre-training from general domain to this domain.

\subsubsection{Results on Text-Based Molecule Generation}
Table \ref{tab:cap2smi} presents the results of different models for molecule generation. Ada-T5 achieves the best performance in all three datasets across almost all metrics, demonstrating its capability to generate high-quality SMILES. The only exception is that the MolXPT slightly surpasses Ada-T5 by 0.009 in ChEBI-20 dataset on the validity metric, which is calculated using RDkit to simply check whether the string can be successfully converted to a molecule object without errors and whether the molecule represents a realistic and feasible chemical structure, without any comparison to the targeted SMILES and the input descriptions. Despite this one slight superiority, MolXPT performs significantly worse than Ada-T5 on other metrics, meaning that although it can generate slightly more valid SMILES, it does not take into account the designated instructions, ergo making it one step away from real-world application. 

On the other hand, Aug-T5 surpasses some existing methods in certain datasets on specific metrics. However, its consistency falls short compared to Ada-T5. This variability may be traced back to the construction of molecule-description data pairs in pseudo data: the LLMS use the real SMILES are used as the input, leaving only the description part of the pseudo data genuinely ``pseudo''. This means that when training Aug-T5 on molecule captioning, it gets the authentic SMILES; but when training on molecule generation, it gets the pseudo description. Consequently, the gap between the input training data leads to the gap between the model performance on different tasks. Furthermore, compared with the results for molecular captioning, the base counterparts of both methods for molecule generation exhibit pronounced enhancements, which could also attributed to the gap between the input data, as using the "pseudo" part as the input for molecule generation might offer more space for improvements, especially for larger-scale models that can better tolerate the ``pseudo" data nuances. 

The difference between Aug-T5 and Ada-T5 also indicates the importance of data authenticity and the difference between real data and pseudo data: as Ada-T5 is later fine-tuned with 100\% real data (in comparison with Aug-T5, which is fine-tuned with the mix of real data and pseudo data), its misunderstandings about SMILES during domain adaptation through pseudo data are corrected and therefore has a better overall performance. This further stresses that using pseudo data for direct application may not be the optimal way to exploit its potential. 

\subsection{Effect of the Amount of Pseudo Data}
In order to further demonstrate how the amount of pseudo data could affect model performance, we experiment on ChEBI-20, the largest and most widely used dataset, with varying numbers of pseudo data samples $\mathcal{N}$ from 1k to 512k. 

\subsubsection{Results on Molecular Captioning}
Figure \ref{fig:ada_aug_smi2cap} shows the results of Ada-T5 and Aug-T5 for molecular captioning with different amounts of pseudo data. Both Ada-T5 and Aug-T5 exhibit significant improvements when a modest amount of pseudo data is incorporated into their training. With just 1k pseudo data, both methods can surpass T5\textsubscript{large} and ChatGPT and achieve a comparative performance to MolT5\textsubscript{large} and MolXPT. This phenomena is often seen in other data augmentation strategies \cite{wei2019eda, sennrich2016improving}, and can be attributed to the moderate noise introduced by the pseudo data, which in turn bolsters model generalization. As the amount of pseudo data increases, Ada-T5 and Aug-T5 exhibit different tendencies. The performance of Aug-T5 begins to decline when the number of pseudo data samples reaches 4k, and sees a sharp drop when it exceeds 32k. This is possibly due to the imbalance between real data and pseudo data: As the model becomes increasingly exposed to unreal patterns from the pseudo data, it might shift its attention away from genuine patterns. Consequently, the real patterns are overlooked by the model that focuses on the artificial ones. In contrast, Ada-T5 thrives with the increasing amount of pseudo data, evidenced by the growth of overall metrics. One possible explanation is that Ada-T5 only uses pseudo data for pre-training, with follow-up fine-tuning using real data. Thus, the increase of pseudo data does not twist its grasp of genuine patterns, but instead, further amplifies the proficiency of the model during subsequent training.

\subsubsection{Results on Text-Based Molecule Generation}
Figure \ref{fig:ada_aug_cap2smi} shows the results of Ada-T5 and Aug-T5 for molecule generation with different amounts of pseudo data.
Ada-T5 shows the same superiority and trend as it does in molecular captioning with more pseudo data incorporated, while Aug-T5 displays a non-linear trend, with the optimal choice of the amount of pseudo data significantly larger than when applying Aug-T5 for molecular captioning. The reason might lie in the dual nature of pseudo data: it introduces both linguistic patterns and noise. Initially, a little bit of pseudo data bolsters model generalization by acting as a regularizer. But as more is added, an overbundance of noise degrades the results. However, once a critical mass of pseudo data is reached, the model starts to recognize more subtle and broader linguistic patterns amidst the noise, which helps in generating more accurate SMILES strings, leading to the observed spike in performance. After this peak, the overwhelming volume of pseudo data might reintroduce the dominance of noise, causing a decrease in performance.

The distinct behavior of Aug-T5 in molecular captioning versus molecule generation highlights their inherent differences. Molecular captioning, being more flexible, can buffer linguistic variations, downplaying minor gains from pseudo data and instead more affected by noise. In contrast, molecule generation requires recognizing specific linguistic cues from descriptions that lead to exact structural changes in the SMILES output, making it more receptive to the subtle intricacies but can also discern and benefit from the subtle patterns present in pseudo data. Overall, these results indicate that the impact of pseudo data varies, depending on its inherent nature and the specific task at hand.

%
%
\section{Conclusion}
In this paper, we introduce a novel approach that enhances low-resource cross-modal molecule discovery by leveraging artificially-real data generated by LLMs. By incorporating a retrieval-based few-shot prompting strategy, we are able to produce high-quality pseudo molecule-description pairs. To mitigate the scarcity of data, we released two datasets: PseudoMD-1M, the first artificially-real dataset for molecule description, and DrugBank-23, a real molecule-description dataset constructed from a novel source. We propose to use pseudo data for domain adaptation and for data augmentation to explore its optimal utilization. Experiments across different datasets show that the former can best exploit the potential of pseudo data, achieving better performance with less parameters and training data. Furthermore, as the performance of the model continues to benefit from the increasing amount of pseudo data, our approach shows the great potential of pseudo data, thereby providing a novel and promising approach for addressing low-resource challenge in cross-modal molecule discovery.

\setcounter{secnumdepth}{0}

\section{Acknowledgements}
We express our gratitude to the anonymous reviewers for their valuable feedback. This research was supported by the National Key R\&D Program of China (2021ZD0113302), the National Natural Science Foundation of China Youth Fund (62206079), and the Heilongjiang Provincial Natural Science Foundation of China (YQ2022F006). We also appreciate Du Xiaoman Technology's support for our research.

\bibliography{ref}

\end{document}